# Advances in deep learning methods for pavement surface crack detection and identification with visible light visual images

LU Kailiang
Jiangsu JARI Technology Group Shanghai Branch
Shengxia Road NO.666, Pudong, Shanghai, P.R. China
lukailiang@163.com

## Abstract

*Cracks inevitably exist widely in buildings, structures, parts or products. Compared to contact detection methods such as nondestructive test (NDT) and health monitoring, surface crack detection or identification with visible light visual images is a kind of non-contact method, which is not limited by the material of the tested object and is easy to achieve online real-time fully automation, thus, has the advantages of fast detection speed, low cost and high precision. Firstly, typical pavement (concrete also) crack public data sets for classification, location or segmentation were collected, and the characteristics of sample images and the random variable factors, including environmental, noise and interference, were summarized. Subsequently, the advantages and shortcomings of the three main crack identification methods, i.e., Hand-crafted Feature Engineering, Machine Learning, Deep Learning, were compared. Finally, from the aspects of model architecture, testing performance and predicting effectiveness, the development and progress of typical deep learning models, namely self-built CNN, transfer learning (TL) and encoder-decoder (ED), which can be easily deployed on embedded platform, were reviewed. Meanwhile, from this, we can see the evolution of CNN model architecture, as well as the obvious improvement of performance and effect because of computing power enhancement and algorithm optimization. The benchmark test shows that: 1) It has been able to realize real-time pixel-level crack identification on embedded platform: the entire crack detection average time cost of an image sample is less than 100ms, either using the ED method (i.e., FPCNet) or the TL method based on InceptionV3. It can be reduced to less than 10ms with TL method based on MobileNet (a lightweight backbone base network). 2) In terms of accuracy, it can reach over 99.8% on CCIC which is easily identified by human eyes. On SDNET2018, some samples of which are difficult to be identified, FPCNet can reach 97.5%, while TL method is close to 96.1%. To the best of our knowledge, this paper for the first time comprehensively summarizes the pavement crack public data sets, and the performance and effectiveness of surface crack detection and identification deep learning methods for embedded platform, are reviewed and evaluated.*

0. Introduction

Structural cracks originate from internal microscopic defects inherent in the material, and are generated and expanded under the action of external forces. As a result, cracks inevitably exist in all kinds of constructions, structures, parts or products along with the extension of their service life in the field of civil construction, transportation, construction machinery and other industries [1-10].

Crack detection or monitoring technology in engineering structures can mainly be divided into:

(1) Nondestructive testing (NDT) techniques [1-2], including radiographic detection (such as CT), ultrasonic detection (such as acoustic emission [3]), electromagnetic eddy current detection, magnetic particle detection and penetrant detection, etc. NDT is applicable to all scenarios of surface, near surface and internal cracks with high measurement accuracy due to its many optional physical means. But it is an active detection technology that requires transmitting and receiving physical medium and may also depend on coupling medium. Therefore, the corresponding detection equipment is usually more complex and expensive.

(2) Structural health monitoring techniques [4], including stress-strain signal based method, vibration signal based method, etc. With the help of analyzing the time history signal collected by pre-arranged sensors, it can identify whether structure has cracks or not. However， it is an indirect measurement method, the identification and positioning accuracy of internal cracks still need to be improved.

(3) Non-contact detection techniques [1-2,5-10,13-20], including infrared detection, laser ultrasonic detection and detection based on visible light visual image. Compared with the former two ways, non-contact detection does not require direct contact with the tested object, thus has the advantages of not being limited by the material of tested object, fast detection speed, easy to realize automatic on-line real-time detection. The disadvantage is that it can be only used for surface crack detection at present.

Surface crack recognition based on 2D visual image has all the advantages of non-contact detection. The physical medium is visible light, with no need of emission source. When the visibility is poor, only supplementary light source is needed. It is almost a passive detection method; the cost advantage of the whole detection system is obvious.

Via Hand-crafted Feature Engineering, Machine Learning and Deep Learning methods, image-based surface crack identification can achieve objectives of qualitative, positional and quantitative crack detection: 1) Classification, to judge whether there is a crack or not; 2) Detection/Location, to detect the location of cracks; 3) Segmentation, to identify the detail of cracks, such as size, topology, etc. Therefore, the surface crack detection and identification technology based on visible light visual images is expected to be applicable to preventive testing or monitoring scenarios such as routine patrol inspection, defect detection of objects that their geometric topological form can be abstracted to 2D plane, where the advantages of fast detection speed, low cost, high precision and easy to achieve online real-time automation of this non-contact method, can be fully utilized.

Taking pavement cracks (including bridge deck, wall concrete cracks as well) as an example, firstly, typical public data sets are collected and the characteristics of sample images are summarized, which is the basis of various feature engineering and learning methods based on statistical theories. Secondly, crack detection or identification algorithms in the three categories of hand-crafted feature engineering, machine learning and deep learning are summarized and compared. Finally, the performance and effectiveness of surface crack detection and identification deep learning models, especially those can be deployed on embedded platform, are reviewed and evaluated.

# 1. Crack public data sets

## 1.1. An overview

We collected public data sets on pavement crack and also concrete crack in the field of civil and construction engineering. The common crack public data sets are shown in Table 1, which mainly lists basic features (e.g., sample number, resolution, colored, and precision of annotation), advanced features (e.g., background, environmental influences, and other generalization factors), license, open sources, etc.

These sample images were all captured by personal cameras, mobile phone cameras or car cameras, thus acquisition cost is low. The number of samples varies from dozens to thousands depending on the purpose and requirement of the data set, and the resolution varies with acquisition device. According to the needs of subsequent processing, original images can be cropped into small samples, such as 256x256 or 227x227 region/patch commonly used in image recognition deep learning algorithm input. The sample number can also be further expanded through data enhancement. In general, compared to ImageNet, Pascal VOC and other general data sets, the collected pavement and concrete crack public data sets belong to special small data sets.

Depending on whether the data set is detection/location oriented or classification/segmentation oriented, sample's crack annotation can be categorized into three precision levels of bounding box annotation, patch/region-level annotation and pixel-level annotation. The crack data sets introduced in Section 1.2 correspond to these three categories respectively. The random variables contained in the sample images were summarized in Section 1.3.

## 1.2. Typical crack data sets

### 1.2.1. JapanRoad

The samples in the JapanRoad data set were taken from real Road street view perspective images captured from the front windshields of cars. They were used for the classification and detection of eight types of road defects, including cracks, as shown in Figure 1. It contains 163,664 real street road images, 9,053 of which are positive samples (with cracks). Bounding box is used for crack location annotation, the same format like PASCAL VOC.

### 1.2.2. SDNET2018

SDNET2018 is a data set with patch/region-level annotation for training, validation and testing artificial intelligence algorithms for concrete crack detection.

The original images were taken with a 16 MP Nikon digital camera. It includes 230 photos of cracked and uncracked concrete surfaces (54 bridge decks, 72 walls, and 104 sidewalks). The deck is located at SMASH LABS in Utah State University; The walls examined belong to the Russell/Wanlass Performance Hall building; Pavement images were taken from the roads and sidewalks of the campus.

Each photo is then clipped and divided into 256×256-pixel samples. If there is a crack in the sample, it is marked as C; if no crack, marked as U. In total, SDNET2018 contains more than 56,000 sample images of cracked and uncracked concrete bridge decks, walls, and pavements. The crack size in the positive samples is as narrow to 0.06 mm and as wide to 25 mm. The dataset also includes image samples with a variety of obstacles and disturbances, including shadows, surface roughness, scaling, edges, holes, and background fragments (as shown in Table 2).

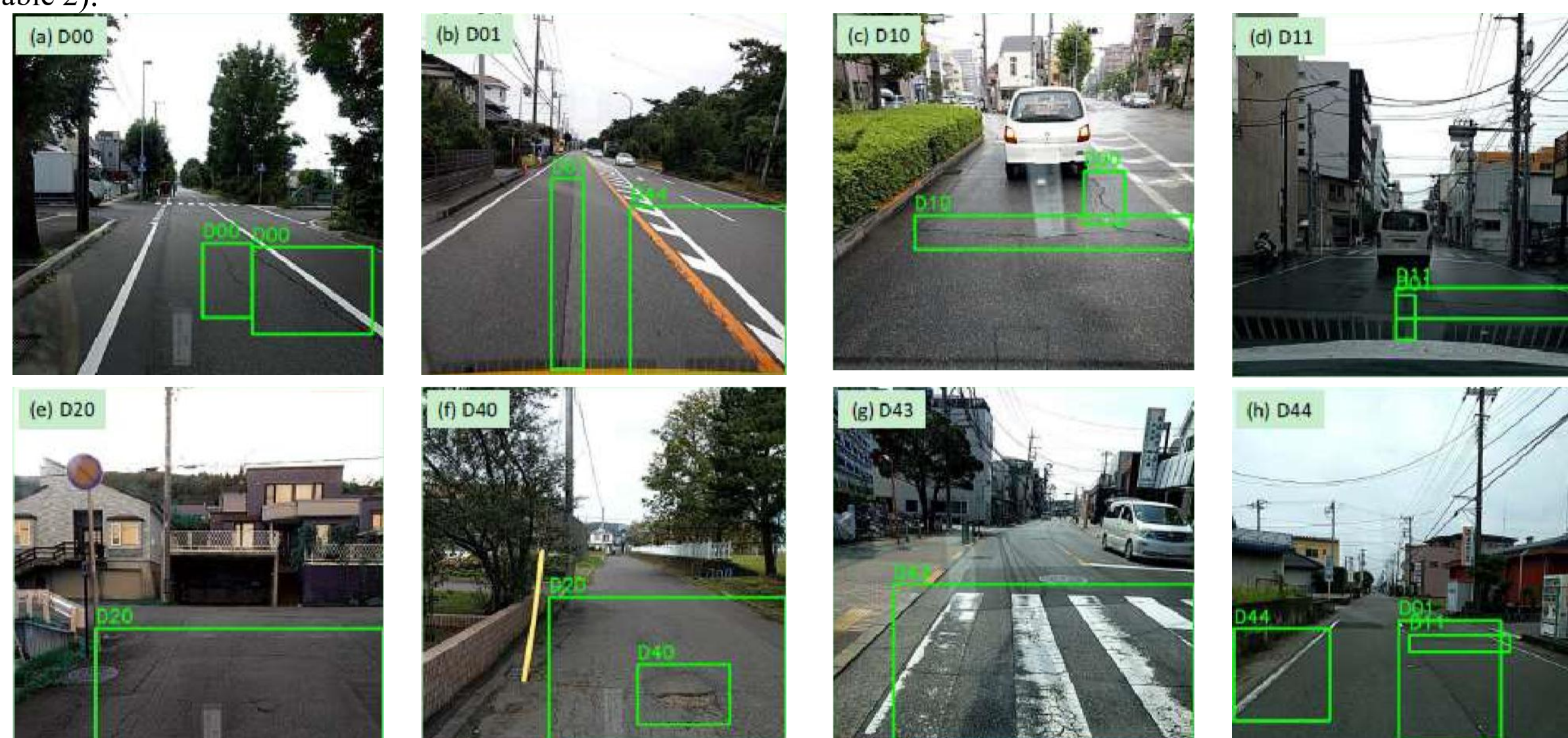

Classes description: (a)D00: Liner crack, longitudinal, wheel mark part; (b)D01: Liner crack, longitudinal, construction joint part; (c)D10: Liner crack, lateral, equal interval; (d)D11: Liner crack, lateral, construction joint part; (e)D20: Alligator crack, Partial pavement, overall pavement; (f)D40: Corruption, Rutting, bump, pothole, separation; (g)D43: Corruption, White line blur; (h)D44: Corruption, Cross walk blur.

**Figure 1 Sample images in JapanRoad data set**

1.2.3. Crack Forest Dataset

Cracked Forest Dataset (CFD) is a pixel-level annotated pavement Crack Dataset reflecting the overall situation of Beijing urban pavement. It is one of the benchmark baseline data sets. Only non-commercial research purposes are currently authorized. A total of 118 images with 480×320 resolution was collected, using a mobile phone (iPhone 5). The images contain noise or interference factors e.g., lane lines, shadows, oil stains, etc. (as shown in Table 2).

**Table 1 Typical crack public data sets**

| Data sets | | Basic features | | | | Advanced features | | Remarks (license, other features) |
|---|---|---|---|---|---|---|---|---|
| category | Name | Image/ Sample NO. | Resolution | Colored | Crack Annotation Level | Background (e.g., car/house /sky) | Environmental or other interfering factors (e.g., shadow/occlusion/low contrast/noise) | |
| Pavement Crack | Crack Forest Dataset(CFD)[1] | 118 | 480×320 | Yes | Pixel-level | Yes \| a little | Lane lines, shadows, noise like oil stain | ·for Non-commercial research purposes only ·one of the benchmark baseline data sets |
| | AigleRN[2] | 38 | 991×462 311×462 | No | Pixel-level | No | Pre-processed to reduce the nonuniform illumination | ·with more complex texture than CFD ·Similar small data sets like ESAR and LCMS |
| | Crack500[3] | 500 | 2000×1500 | Yes | Pixel-level | No | Lane lines, oil/wet stain, tire brake marks, speckle noise, etc. | ·Covers almost all kinds of features except for shadow |
| | GAPs[3] | 1969 | 1920×1080 | No | Pixel-level | No | Little noise like oil stain, asphalt and rails, lane lines | / |
| | Cracktree200[3] | 206 | 800×600 | Yes | Pixel-level | No | only shadow & lane lines | / |
| | G45[4] | 122 | 2048×1536 | No | Pixel-level | No | Brightness change, oil | ·4 types: transverse, |

[1] https://github.com/cuilimeng/CrackForest-dataset
[2] https://www.irit.fr/~Sylvie.Chambon/Crack_Detection_Data
[3] https://github.com/fyangneil/pavement-crack-detection
[4] https://github.com/YuchunHuang/FPCNet

| | | | | | | | | |
|---|---|---|---|---|---|---|---|---|
| | | | | | | | stain, tire brake marks, speckle noise, etc. | longitudinal, block, and alligator cracks |
| | EdmCrack600[5] | 600 | 1920×1080 | Yes | Pixel-level | Yes | Weather, illumination, shadow, texture difference | ·only for academic research<br>·Perspective image taken from the rear camera of a car |
| | JapanRoad[6] | 9053 | 600×600 | Yes | Bounding Box | Yes | Real street view perspective images captured from the front windshields of a car | ·License: CC BY-SA 4.0<br>·for road defect detection<br>·PASCAL VOC annotation format |
| Concrete Crack | SDNET2018[7] | 230 (cropped to 56092) | 4096x3840 (256×256) | Yes | Patch-level | Yes \| various obstacles | with a variety of obstructions, including shadows, surface roughness, scaling, edges, holes, and background debris | ·License: Creative Commons Attribution 4.0<br>·54 bridge decks, 72 walls, 104 pavements<br>·Crack size: 0.06-25mm |
| | Concrete Crack Images for Classification[8] | 458(cropped to 40000) | 4032×3024 (227×227) | Yes | Patch-level | No | Shadows, lighting spot, blurred, including thin & close-up cracks | ·Similar to SDNET2018, images collected from METU Campus Buildings |

**Table 2 Environmental influence factors in crack data sets**

| Environmental factors | Images (positive samples with crack) |
|---|---|
| No | Aigle-RN |
| Illumination change<br>Shadow<br>Lighting spot<br>low contrast<br>etc. | CFD Cracktree200<br>SDNET2018 |

**Table 3 Noise or interference factors in crack data sets**

| Noise or interference | | Images | |
|---|---|---|---|
| Data set | Factors | Negative samples without crack | Positive samples with crack |
| CFD | Lane lines, oil stain, manhole cover | | |

[5] https://github.com/mqp2259/EdmCrack600
[6] https://github.com/sekilab/RoadDamageDetector/
[7] https://digitalcommons.usu.edu/all_datasets/48/
[8] https://data.mendeley.com/datasets/5y9wdsg2zt/1

| | | | |
|---|---|---|---|
| GAPs | Asphalt, lane lines, rails, manhole cover | | |
| Crack500 | Wet stain, spots | | |
| CCIC (most samples can be easily identified by human eyes) | Blurred, lighting spot/spots, texture | | |
| SDNET2018 (some samples are difficult to be identified by human eyes) | Oil stain, blurred, edge | | |
| | Texture, speckle, surface roughness | | |

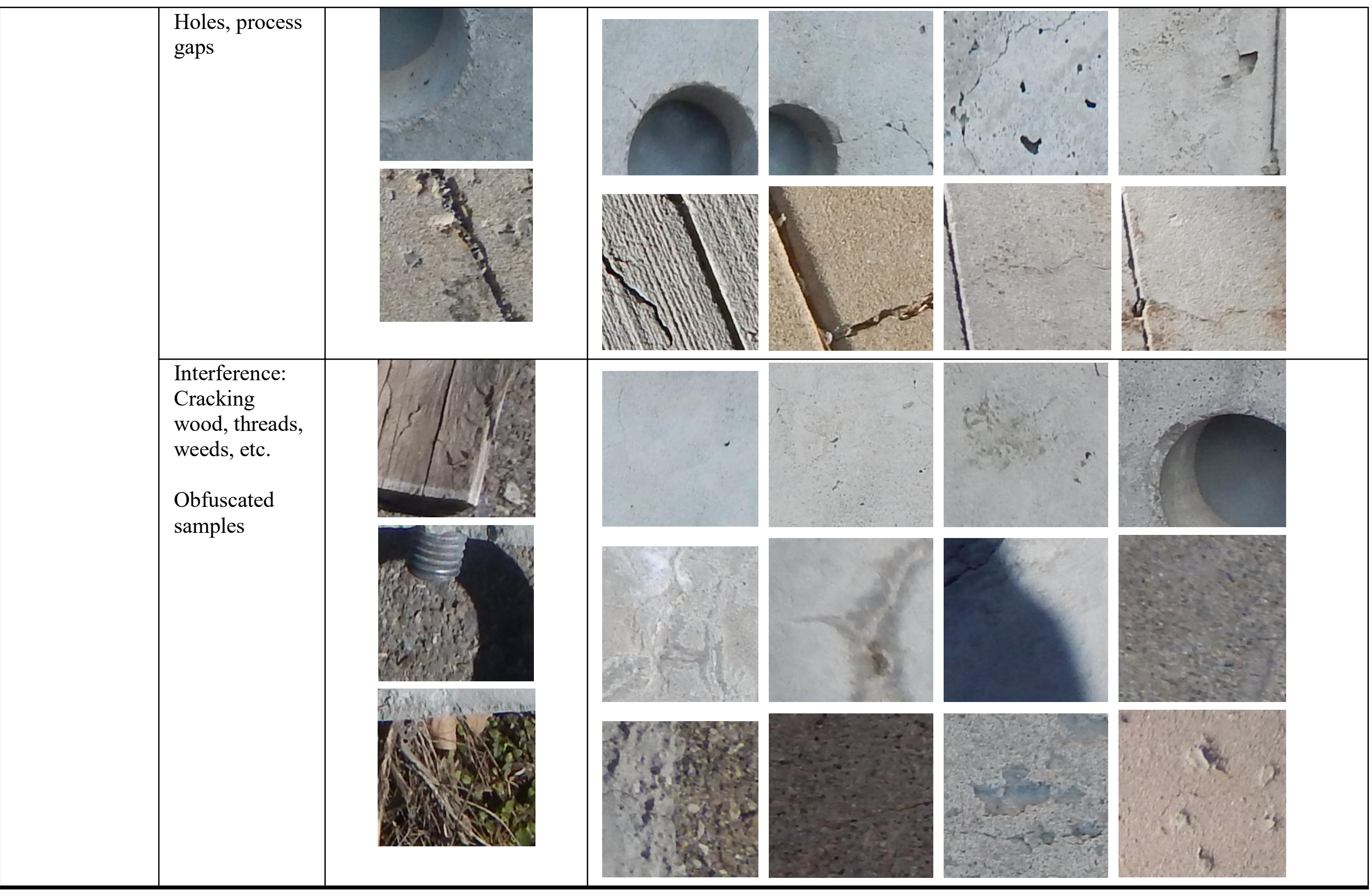

### 1.3. Random variable influencing factors in image samples

The more random variable influencing factors involved in a data set, the more advanced features it contains, and then the generalization ability of the machine learning/deep learning model trained and validated by the data set is stronger in real test scenario. These random variable factors mainly include background, environmental and other interference factors, and are combined and superimposed in sample images. For the sake of clarity and presentation, the sample images shown below often illustrate only a single factor.

#### 1.3.1. Background

Background of the road real scene acquisition images is rich, as shown in Figure 1. In JapanRoad, the background includes cars, houses, sky, pedestrians, green belt, lane lines, telegraph poles, etc. CFD also contains a small number of samples with street background. Another kind of background is noise or interference, such as oil stain, tire brake mark, spots, see details in Section 1.3.3.

#### 1.3.2. Environmental influencing factor

Environmental influencing factors mainly refer to brightness variation, shadow, lighting spot and low contrast caused by weather or illumination changes. As shown in Table 2, in contrast, some samples of Aigle-RN that were pre-processed specifically to reduce uneven lighting influence on the complex crack texture, are presented for comparison.

#### 1.3.3. Noise or interference factors

There are other noise or interference factors such as lane lines, oil/wet/ asphalt stain, spots, manhole covers, tire brake marks, texture difference, surface roughness, boundaries/edges, holes and other sundries, as shown in Table 3.

## 2. Crack detection and identification methods comparison

According to the general classification of image identification methods, crack recognition algorithms can be divided into: hand-crafted feature engineering, machine learning and deep learning. The common algorithms and their characteristics of the three types are shown in Table 4.

### 2.1. Hand-crafted feature engineering

Hand-crafted feature engineering method usually includes edge/morphology/feature detection algorithms and feature transformation (or filtering) algorithms. For example, Canny [7-8], Sobel [16], Histogram of Oriented

Gradient (HOG), Local Binary Pattern (LBP), belong to the former, fast Fourier Transform (FFT), fast Haar Transform (FHT), Gabor filter, intensity thresholding [8] are the latter ones.

These algorithms gradually extract or analyze the edge, morphology and other features of objects in image by means of mathematical calculation. They are not learning methods and do not depend on data set. And the mathematical calculation is mostly analytical formula, generally speaking, is light in computation and fast. The shortcomings of these algorithms is the adaptive to various random variable factors.

## 2.2. Machine learning

Both machine learning and deep learning are learning methods and depend on data set. The difference is that the mathematical expression of the former is explicit and explicable, while that of the latter is implicit. It reaches the consensus that deep learning (the depth of neural network is deeper) is oriented towards higher-dimensional features, while machine learning is oriented towards lower-dimensional features. Therefore, machine learning is somewhere between hand-crafted feature engineering and deep learning methods.

Support Vector Machine (SVM) [7], Decision Tree and Random Forest are common machine learning algorithms. Correspondingly, crack classification and identification algorithms mainly include CrackIT, CrackTree, CrackForest [8,10,16-19] etc.

## 2.3. Deep learning

Thanks to 1) parallel computing hardware (e.g., GPU/TPU), 2) large data sets (e.g., ImageNet, Pascal VOC) as the benchmark baseline, and 3) algorithm is continuously improved and perfected, such as Batch Normalization, Residual Connection and Depth of Separable Convolution; deep learning has gained prominence since 2012. However, the mathematical theory of deep learning method is not complete, that is, the "black box" and interpretability problem. Nevertheless, it is undeniable that it has made SOTA(State-of-the-Art) progress and even surpassed the human level in computer vision (CV), natural language processing (NLP) and reinforcement learning. At present, scientists are trying to solve theoretical problems [11], and engineers are also improving the interpretability of the deep learning process through visualization technology [12].

The pattern learned by Convolutional Neural Network (CNN) is translation invariant and has spatial hierarchies, which are also the core features of higher animal (e.g., cats, human) vision. As a result, breakthroughs have been made in the field of computer vision identification since 1998, especially since 2012, with algorithms by using these two features.

Image identification algorithms based on CNN have the advantages of automatic feature extraction, strong generalization ability, high precision. In these respects, it is better than machine learning and hand-crafted feature

**Table 4 Common algorithms and their characteristics for crack identification**

<table>
<tr><th colspan="2">Method</th><th>Algorithms</th><th colspan="2">Characteristics</th></tr>
<tr><td colspan="2">Hand-crafted Feature Engineering</td><td>Edge/morphology/feature detection algorithms: Canny, Sobel, Histogram of Oriented Gradient (HOG), Local Binary Pattern (LBP), etc.<br>Feature transformation algorithms: fast Haar transform (FHT), fast Fourier transform (FFT), Gabor filters, Intensity thresholding, etc.</td><td colspan="2">√Don't have to learn from data set<br>√Usually, light computation and fast<br>·Weak generalization ability to various random variable factors. Once the application scenario or environment changes, parameters need to be fine-tuning, or the algorithm needs to be redesigned and even fails at all</td></tr>
<tr><td colspan="2">Machine Learning</td><td>CrackIT, CrackTree, CrackForest, etc.</td><td colspan="2">·Learning methods for low-dimensional features, between hand-crafted feature engineering and deep learning, and need data set.</td></tr>
<tr><td rowspan="5">Deep Learning</td><td>Self-built Architecture</td><td>Self-built CNN [7], Structured-Prediction CNN [8], CrackNet/CrackNet-V [9-10]</td><td rowspan="5">√Automatic feature extraction<br>√Strong generalization ability<br>√High precision<br>·Need data set<br>·The "black box" problem</td><td>√Be designed for specific data sets and application scenarios<br>√Lightweight model and few parameters<br>·not SOTA model, generalization ability is not so good</td></tr>
<tr><td>Transfer Learning, TL</td><td>VGG-16 based TL [13], Inception-v3 based TL [14], Xception based TL [15], VGG-19/Resnet152 based TL [16]</td><td>√Fine-tuning on SOTA models[9] that have been architecture-optimized and pre-trained, easy to adjust, good generalization performance and the training is fast</td></tr>
<tr><td>Encoder-Decoder, ED</td><td>FPCNet: Fast Pavement Crack Detection Network [17]</td><td>√Suitable for weakly supervised and small data set situations, with excellent accuracy and speed<br>√FPCNet is one of SOTA models</td></tr>
<tr><td>Generative Adversarial Networks, GAN</td><td>CrackGAN: per-pixel semantic segmentation method [18]<br>ConnCrack: combining cWGAN & connectivity maps [16]</td><td>√High accuracy, can used for difficult samples mining<br>·Long training and prediction time (1.56s/img of 1920×1080, @NVIDIA 1080Ti)</td></tr>
<tr><td>Others</td><td>Feature Pyramid and Hierarchical Boosting Network (FPHBN) [19]</td><td>·Poor performance compared to FPCNet [17]</td></tr>
</table>

[9] The evolution of common CNN backbone models seen https://github.com/mikelu-shanghai/Typical-CNN-Model-Evolution

engineering. These algorithms can be divided into different paradigms: 1) Self-built CNN Architecture; 2) Transfer Learning (TL); 3) Encoder-decoder (ED); 4) Generative Adversarial Networks (GAN); etc.

Self-built CNN architecture, such as CrackNet/CrackNet-V [9-10], is a kind of paradigm to design and build model for specific data sets for the application scenarios, according to general design principles of CNN. It has the advantages of good adaptability to specific application scenarios, lightweight model and few parameters, etc. On the other hand, the generalization ability is not so good, and the model is often not optimized thus not SOTA.

Transfer Learning (TL) uses CNN backbone networks to learn basic low-dimensional features. These backbone networks (e.g., VGG-16[13], VGG-19[16], Inception [14], Xception [15], Resnet152[16], seen in Section 3.2.2) are usually SOTA models that have been structure-optimized and pre-trained on large-scale data sets. Then the weight parameters of the upper and/or output layers are fine-tuned on specific data set, to learn higher-dimensional features. TL is applicable to small data set problem, easy to adjust, good generalization performance and the training is fast.

Encoder-decoder (ED) consists of an Encoder and a Decoder. In CNN, the role of encoder network is to produce feature maps with semantic information. The role of the decoder network is to map the low-resolution features output back to the size of the input image. ED is suitable for weakly supervised and small data set. ED based FPCNet [17] is one of SOTA models with excellent accuracy and speed.

Generative Adversarial Networks (GAN) make the generated samples consistent with the probability distribution of real data by means of two networks' confrontation training. One is discriminant network the goal of which is to determine as accurately as possible whether a sample is generated from real data or from a generative network. The other is a generative network, the goal of which is to generate samples that the discriminant network cannot distinguish from real data. The two networks with opposite goals are constantly alternating training. When it finally converges, if the discriminant network can no longer determine the source of a sample, then it is to say that the generative network can generate samples that conform to the real data distribution. CrackGAN [18] and ConnCrack [16] are representative GAN models for crack identification, with high accuracy, and can also be used for difficult samples mining. But the training and prediction time is longer. For example, it is estimated that predicting a 1920×1080 image needs about 1.56s on NVIDIA 1080Ti GPU.

Other algorithms include Feature Pyramid and Hierarchical Boosting Network (FPHBN) [19], which combines feature pyramid and hierarchical boosting. The model architecture is similar to FPCNet, but has poor performance compared to FPCNet.

Section 3 will introduce the development and progress of the deep learning methods including self-built CNN, transfer learning, and encoder-decoder, focusing on the architecture, performance and effectiveness of typical crack identification models. GAN and other algorithms will not be discussed further in this paper due to the performance problem of deployment on current embedded platforms.

## 3. Progress in deep learning method for crack identification

### 3.1. Self-built CNN architecture

Earlier ConvNet [7], Structured-Prediction CNN [8] and recent CrackNet/CrackNet-V [9-10] models were introduced, compared to benchmarks with hand-crafted feature engineering and machine learning methods. From which, the evolution of CNN model architecture, as well as the obvious improvement of performance and effect because of computing power enhancement and algorithm optimization, can be seen.

#### 3.1.1. ConvNet

(1) Data set

It contains 500 original images with a resolution of 3264×2448, cropped into a training set of 640,000 samples, a validation set of 160,000 samples and a testing set of 200,000 samples. All samples are with a resolution of 99×99 and patch-level annotated. The data set is non-public.

(2) Model architecture

The model is relatively simple, including 4 convolutional layers (4×4, 5×5, 3×3, 4×4) and 2 fully-connected layers. There is a max. pooling (mp) layer after each convolutional layer (conv), as shown in Figure 2.

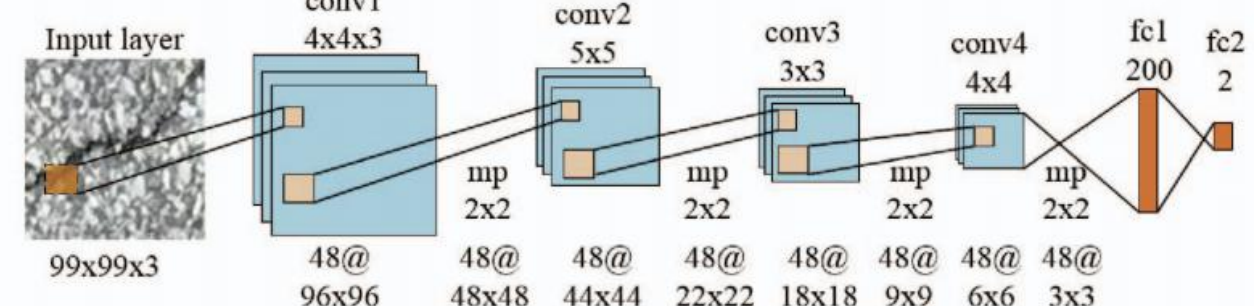

**Figure 2  ConvNet architecture [7]**

(3) Performance metrics

The accuracy metrics of sample test are mainly as follows.

$$Accuracy = \frac{TP+TN}{Total(TP+FN+FP+TN)} \quad (1)$$

$$Precision = \frac{TP}{(TP+FP)} \quad (2)$$

$$Recall = \frac{TP}{(TP+FN)} \quad (3)$$

$$F1\ \ Score = \frac{2 \times Precision \times Recall}{Precision+Recall} = \frac{2 \times TP}{Total+TP-TN} \quad (4)$$

In Equation (1-4), TP denotes True Positive, TN denotes True Negative, FP denotes False Positive, and FN denotes

False Negative. Equation (4) shows that $F1$ *Score* is the weighted harmonic average of *Precision* and *Recall*.

Other measures include the ROC curve etc.

(4) Performance and effect

The performance and effect comparison between ConvNet [7] and SVM, boosting methods are shown in Table 5 and Figure 3. It is not difficult to find out that ConvNet [7] is obviously superior to traditional machine learning methods in both accuracy measurement and recognition effect. Since ConvNet [7] was an early exploratory study of the deep learning method on crack identification, it only focused on the accuracy and did not involve the test speed performance.

**Table 5 Comparison of test performance between ConvNet and SVM, Boosting methods**

| Method | Precision | Recall | *F*1 |
|---|---|---|---|
| SVM | 0.8112 | 0.6734 | 0.7359 |
| Boosting | 0.7360 | 0.7587 | 0.7472 |
| ConvNet [7] | **0.8696** | **0.9251** | **0.8965** |

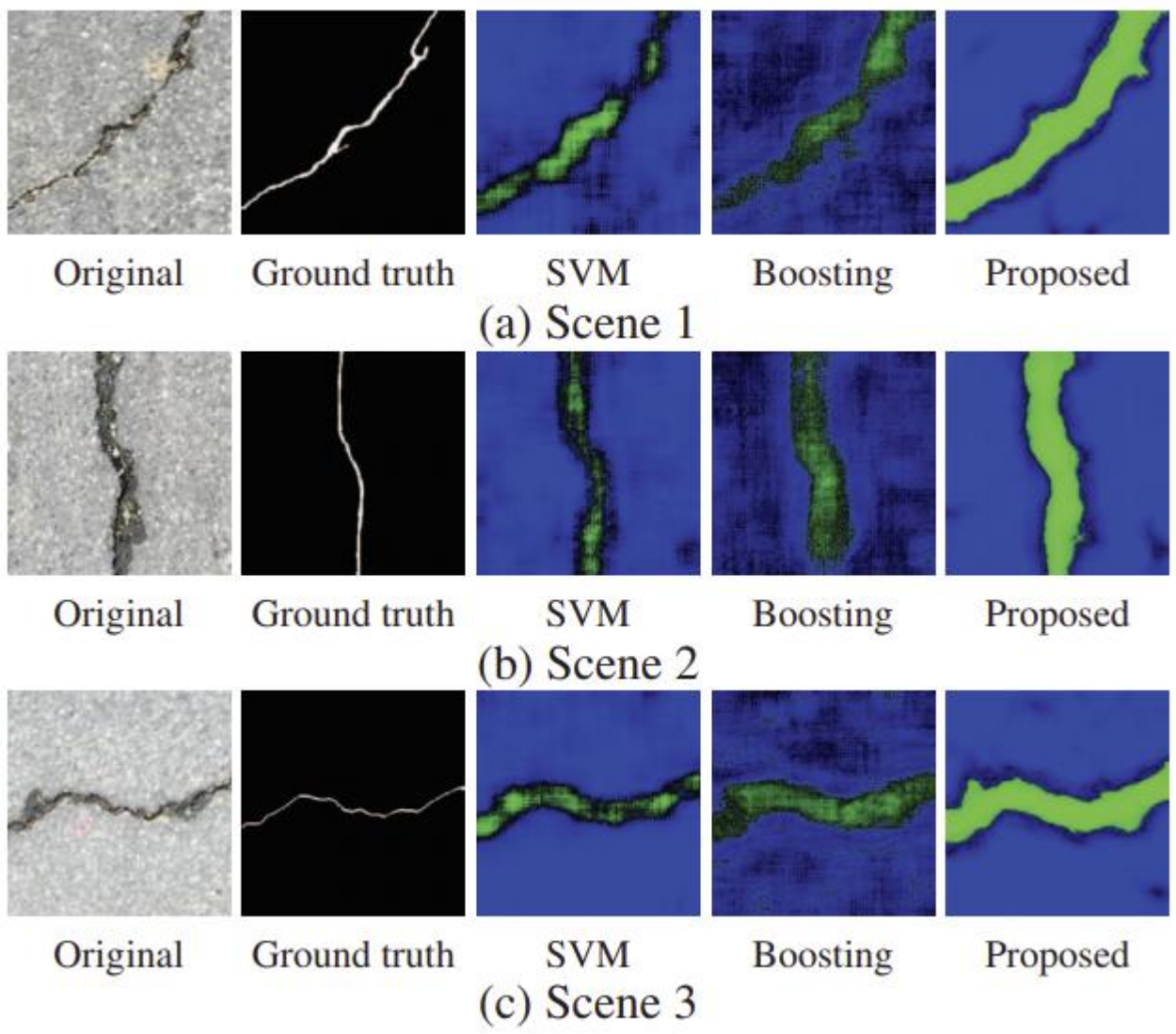

**Figure 3 Probability maps [7]**

3.1.2. Structured-Prediction CNN

(1) Data set

Structure-prediction CNN [8] focused on both test accuracy and speed performance. The model was tested on the open dataset AigleRN to form a benchmark comparison, and an open CFD was also built to be a baseline dataset. The original images in data set were cropped to samples of 27×27 as input.

(2) Model architecture

As shown in Figure 4, the leftmost block is a color image input sample of 27×27 with 3 channels. The other cubic blocks represent feature maps obtained from convolution or max pooling operations. The kernel size of all convolutional layers is 3×3, with a stride of 1 and a padding of 0. The max pooling layers use a 2×2 window with a stride of 2. Two fully-connected layers and one output layer are final ones.

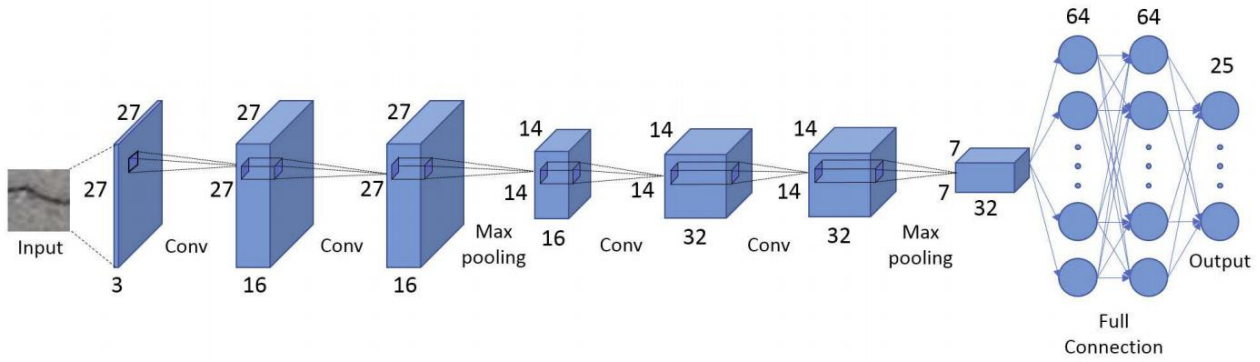

**Figure 4 Structured-Prediction CNN architecture [8]**

(3) Testing performance

In terms of accuracy, the results shown in Table 6 again confirm that deep learning CNN performs better than machine learning method (i.e., CrackForest), and much better than hand-crafted feature engineering method (i.e., Canny, Local Thresholding etc.).

In terms of speed, Structured-Prediction CNN [8] can achieve 2.6 FPS when predicting original images of CFD at NVIDIA GTX1080Ti GPU.

**Table 6 Performance comparison of Crack detection results**

| Method | Precision | Recall | *F*1 |
|---|---|---|---|
| Canny | 0.4377 | 0.7307 | 0.4570 |
| Local thresholding | 0.7727 | 0.8274 | 0.7418 |
| CrackForest | 0.7466 | **0.9514** | 0.8318 |
| Structured-Prediction CNN | **0.9119** | 0.9481 | **0.9244** |

(a) Crack detection results on CFD

| Method | Precision | Recall | *F*1 |
|---|---|---|---|
| Canny | 0.1989 | 0.6753 | 0.2881 |
| Local thresholding | 0.5329 | 0.9345 | 0.6670 |
| FFA | 0.7688 | 0.6812 | 0.6817 |
| MPS | 0.8263 | 0.8410 | 0.8195 |
| Structured-Prediction CNN | **0.9178** | **0.8812** | **0.8954** |

(b) Crack detection results on AigleRN

(4) Identification effect

Structured-Prediction CNN [8] adopted a small-size input of 27×27, moreover, both CFD and AigleRN made pixel-level annotation (2-pixel errors) on the cracks. Thus, the crack detection effect is much better than ConvNet [7], and can be used to identify alligator cracks, as shown in Figure 5.

3.1.3. CrackNet and CrackNet-V

(1) Data set

Different from previous models, CrackNet/CrackNet-V [9-10] used 3D asphalt pavement images obtained by laser scanning as input. Compared with images captured by visible light cameras, laser scanning images have higher resolution and can filter some noise and random factor interference, which is equivalent to an image preprocessing. As a result, the image is clearer and the pixel-level cracks can be perfectly labeled and segmented.

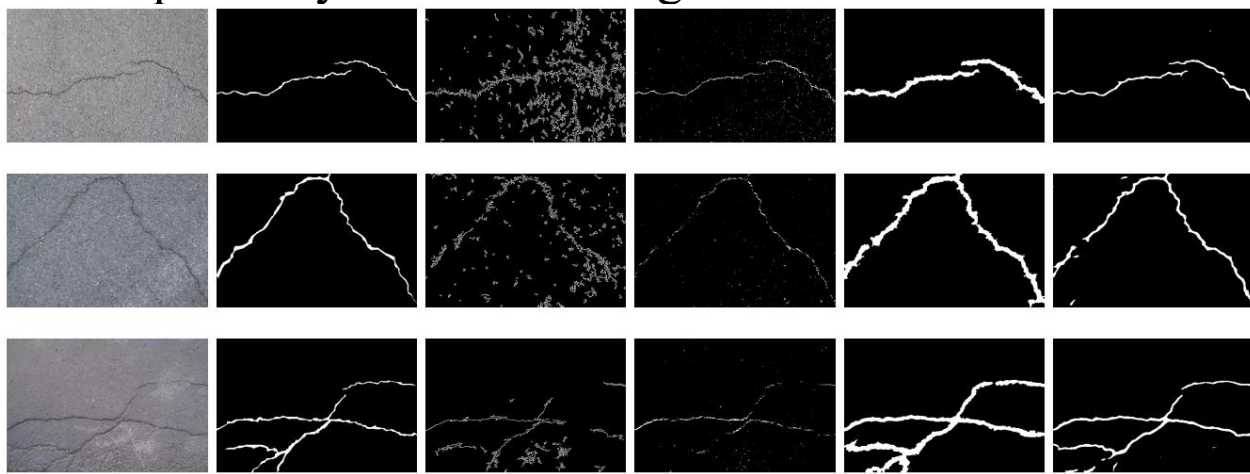

(a) Results comparing on CFD, from left to right: original image, ground truth, Canny, local thresholding, CrackForest, Structured-Prediction CNN

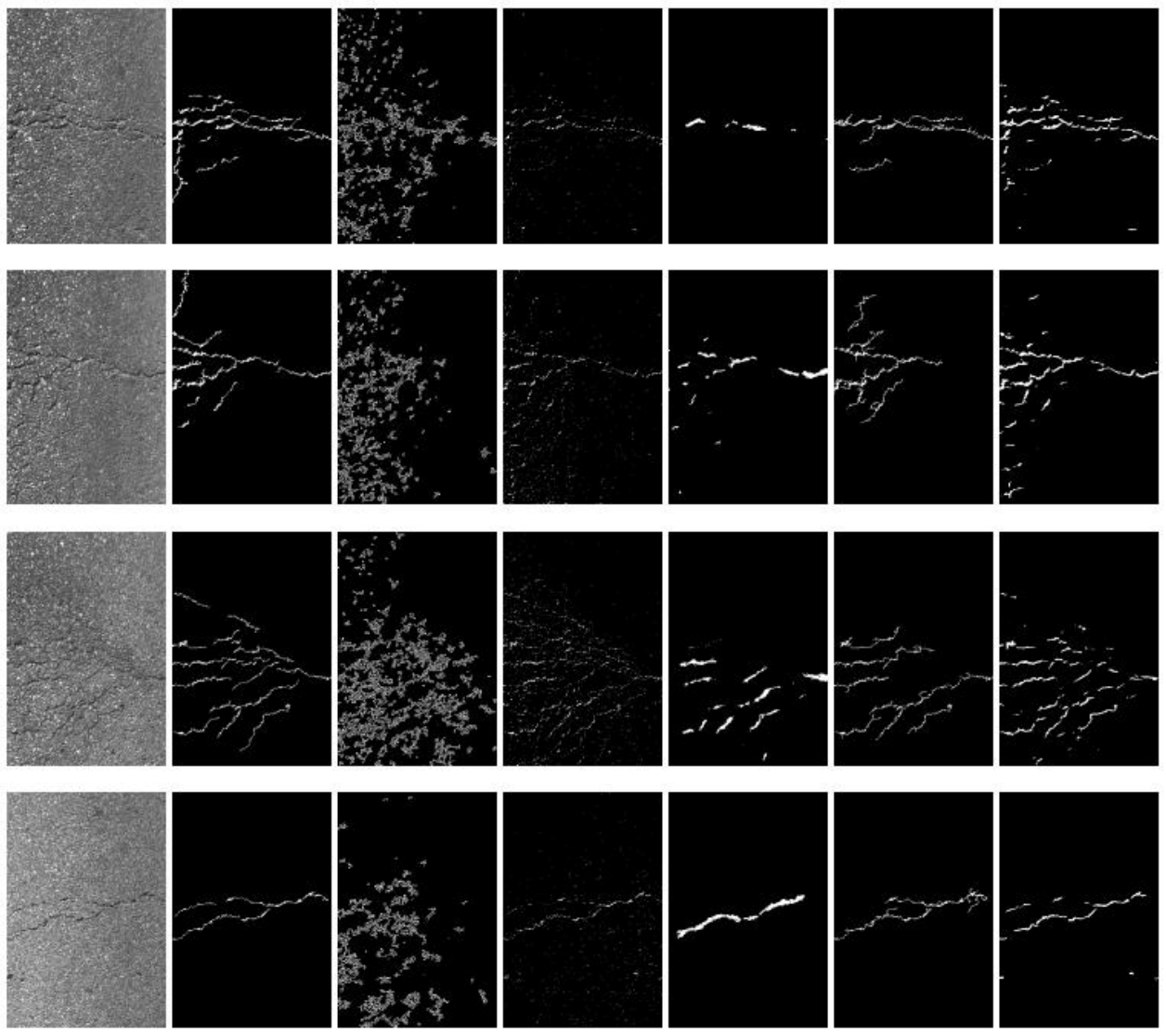

(b) Results of different methods on AigleRN, from left to right: original image, ground truth, Canny, local thresholding, FFA, MPS, Structured-Prediction CNN

**Figure 5 Results comparing of different methods on CFD or AigleRN [8]**

Therefore, not only images or signals collected by cameras, but also by laser scanner, infrared camera and even acoustic emission etc. NDT equipment, can be used as inputs to deep learning models. Thus, not only surface cracks can be identified, but also internal cracks are expected to be detected or monitored. Of course, the emphasis of this paper is to review the advances of surface crack recognition deep learning methods based on visible light visual image.

The 3D asphalt pavement image data set includes a training set of 2,568 image samples, a validation set of 15 typical image samples and a test set of 500 image samples. The input sample resolution of CrackNet [9] is 1024×512, while that of CrackNet-V [10] is 512×256.

(2) Model architecture

The model architecture of CrackNet [9] is shown in Figure 6, including the input layer, two convolutional layers (kernel is 50×50 and 1×1 respectively), two fully-connected layers and the output layer, with a total of 1,159,561 parameters.

The model architecture of CrackNet-V [10] is shown in Figure 7, adding preprocessing layers (median filtering and Z Normalization). 3×3 convolutional layers were widely used, referring to VGG architecture, plus a 15×15 convolutional layer and two 1×1 convolutional layers. It makes CrackNet-V [10] deeper than CrackNet [9] but more lightweight, with only 64,113 parameters, because there is no fully-connected layers and the convolution kernel is much smaller.

Neither CrackNet [9] nor CrackNet-V [10] has max. pooling layer, so the resolution of input and output images remains the same, thus perfect pixel-level segmentation can be achieved.

(3) Testing performance

Comparing Table 7 and Table 8, in terms of accuracy, CrackNet-V [10] is slightly better than CrackNet [9]. In terms of speed, the former is about a quarter of the latter, regardless of the training and testing time (both forward propagation and back propagation), which proves the optimization effect of CrackNet-V [10] architecture.

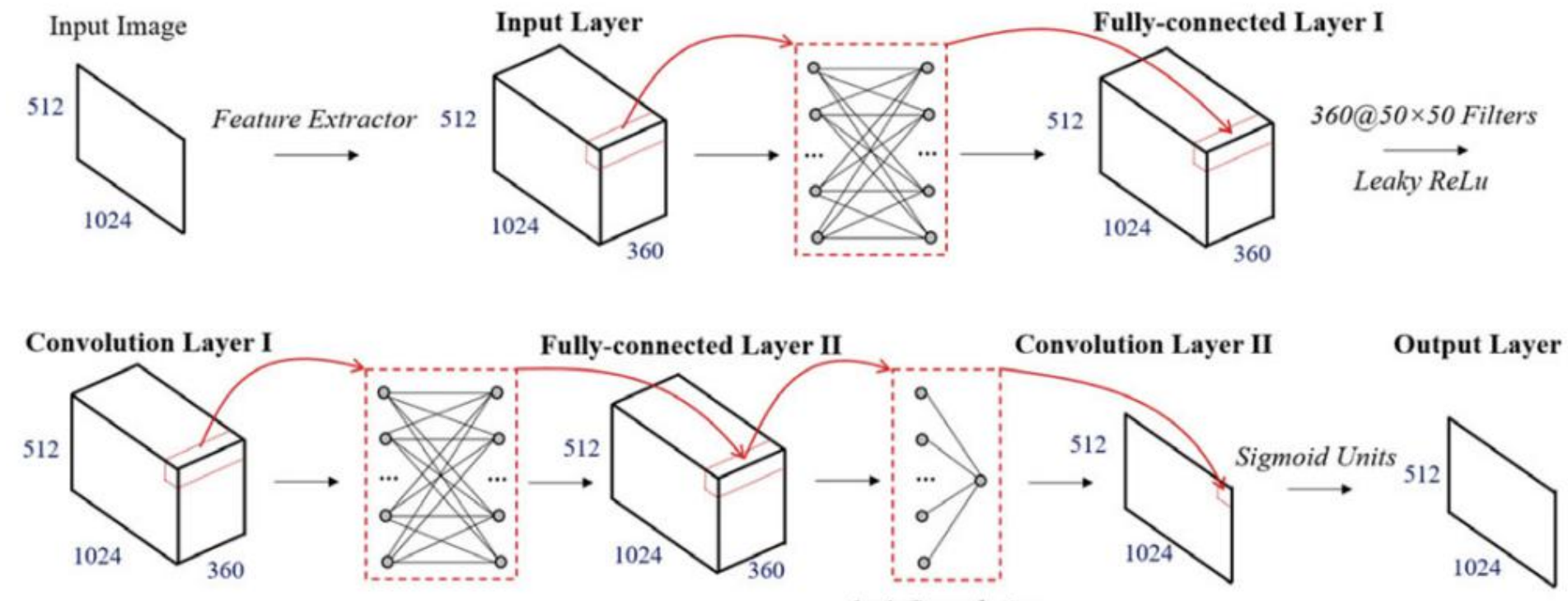

**Figure 6 CrackNet architecture [9]**

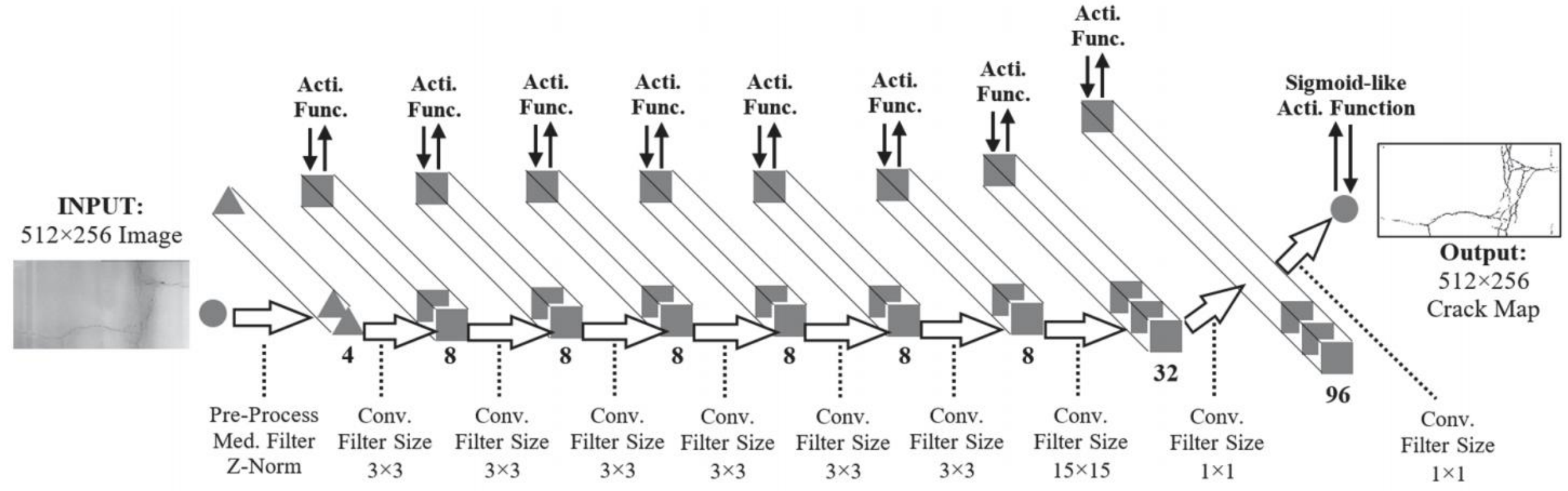

**Figure 7 CrackNet-V architecture [10]**

**Table 7 Comparison between CrackNet and CrackNet-V on testing accuracy**

| Network | No. of parameters | Precision | Recall | $F1$ |
|---|---|---|---|---|
| CrackNet [9] | 1,159,561 | 0.9086 | 0.8096 | 0.8562 |
| CrackNet-V [10] | 64,113 | 0.8431 | 0.9012 | 0.8712 |

**Table 8 Comparison between CrackNet and CrackNet-V on training and testing speed**

| Network | Forward-prop. Speed (s) | Back-prop. Speed (s) | Training time (day) |
|---|---|---|---|
| CrackNet [9] | 1.21 | 8.25 | 4 |
| CrackNet-V [10] | 0.33 | 2.28 | 1 |

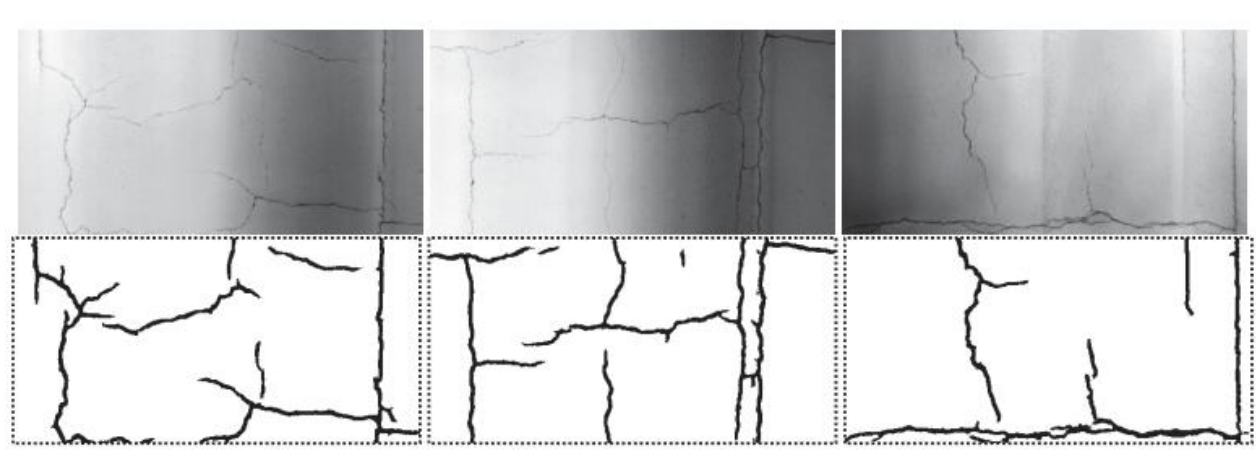

**Figure 8 CrackNet-V crack detection results [10]**

(4) Identification effect

The identification effect of CrackNet-V [10] on alligator cracks is shown in Figure 8. Because of the special optimization design of the model architecture, pixel-perfect crack labeling and segmentation can be achieved. On the other hand, the detection speed is also very fast, which is about 0.33s/img (512×256) on NVIDIA GTX1080Ti GPU. It is almost the same as Structure-prediction CNN [8].

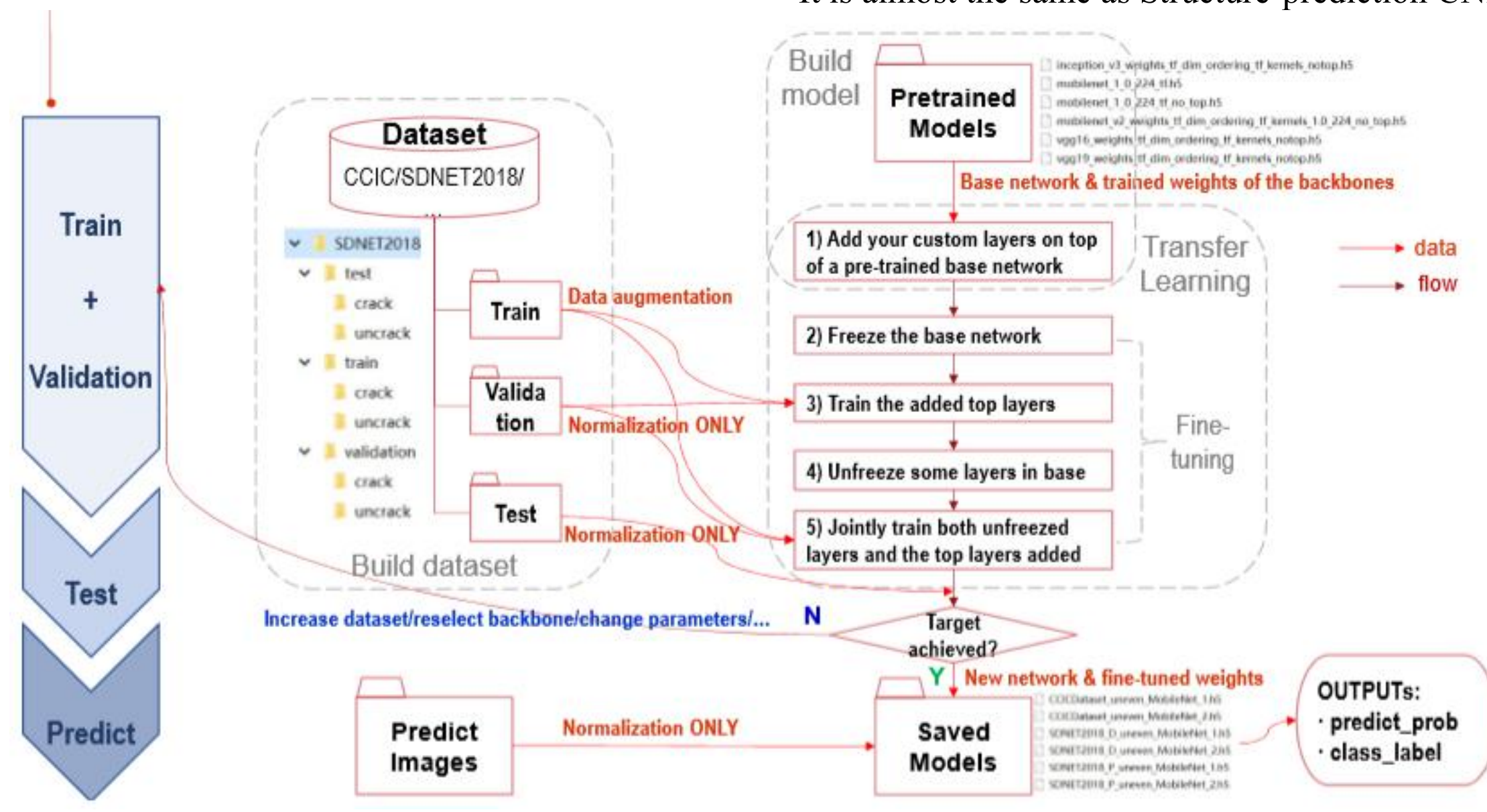

**Figure 9 Framework and steps of transfer learning**

### 3.2. Transfer learning

#### 3.2.1. Framework and steps of transfer learning

The overall procedure of deep learning are training, Validation, testing, and prediction. The framework and steps for transfer learning is shown in Figure 9 in detail.

#### 3.2.2. Comparison of backbone models for transfer learning

The common backbone models for transfer learning are list in Table 9. The memory size, and the accuracy of top-1 and top-5 on ImageNet validation set were compared.

**Table 9 Common backbone models for transfer learning[10]**

| Model | Memory size (MB) | Top-1 Accuracy | Top-5 Accuracy |
|---|---|---|---|
| Xception | 88 | 0.790 | 0.945 |
| VGG16 | 528 | 0.713 | 0.901 |
| VGG19 | 549 | 0.713 | 0.900 |
| ResNet50V2 | 98 | 0.760 | 0.930 |
| ResNet101V2 | 171 | 0.772 | 0.938 |
| ResNet152V2 | 232 | 0.780 | 0.942 |
| InceptionV3 | 92 | 0.779 | 0.937 |
| InceptionResNetV2 | 215 | 0.803 | 0.953 |
| MobileNet | 16 | 0.704 | 0.895 |
| MobileNetV2 | 14 | 0.713 | 0.901 |
| DenseNet121 | 33 | 0.750 | 0.923 |
| DenseNet169 | 57 | 0.762 | 0.932 |
| DenseNet201 | 80 | 0.773 | 0.936 |
| NASNetMobile | 23 | 0.744 | 0.919 |
| NASNetLarge | 343 | 0.825 | 0.960 |

#### 3.2.3. Benchmark test result and performance

Considering the tradeoff between accuracy and efficiency, and prediction deployment on embedded platforms, InceptionV3 and MobileNetV1/V2 were selected for benchmark testing on CCIC and SDNET2018 datasets. The result and performance are compared with models proposed by relevant literatures, as shown in Table 10.

It can be concluded from Table 10 that:

1) Using transfer learning method to classify crack or uncrack images, the accuracy of testing has exceeded the baseline of ImageNet multiple classification (seen in Table 9). On similar comparable data sets (e.g., CCIC), it is also superior to the current published methods [13-14,20]. For example, the testing accuracy of migration learning using lightweight backbone model MobileNetV1 can reach 99.8% on CCIC data set whose samples can be easily recognized by human eyes.

2) The testing accuracy of TL-InceptionV3 on SDNET2018 dataset, some samples of which are difficult to be recognized by human eyes, is 96.1%, that is close to SOTA 97.5% by FPCNet [17]. The accuracy of transfer learning can be further improved with a more heavyweight backbone model. While, predicting time of TL-InceptionV3 cost 16.1ms, much lower than 67.9ms of ED-FPCNet [17], both on NVIDIA GTX1080Ti GPU platform.

### 3.3. Encoder-decoder

There are two shortcomings in the above CNN/FCN methods: 1) There are different widths and topologies in pavement cracks. However, the receptive field of the CNN/FCN filter to extract features is a kernel of specific size (especially the model after VGG mainly adopts 3×3). 2) It is not considered that crack edges, patterns or texture contribute differently to detection results. The Encoder-decoder model FPCNet [17] was developed to solve these deficiencies.

#### 3.3.1. FPCNet

(1) Data set

Tests were carried out on CFD (sample cropped to 288×288) and G45 (sample cropped to 480×480) data sets.

(2) Model architecture

The model architecture of FPCNet [17] includes two modules: Multi-Dilation (MD) module and SE-Upsampling module.

The MD module, as shown in Figure 10, concatenates four dilated convolutions with rates of {1, 2, 3, 4} (seen in Figure 11), a global pooling layer and the original crack multiple-convolution (MC) features. After the concatenation, a 1×1 convolution is performed to obtain the crack MD features. Every convolution retains its number of feature channels except the last 1×1 convolution, and padding is used to ensure that the resolution of the MC feature remains constant.

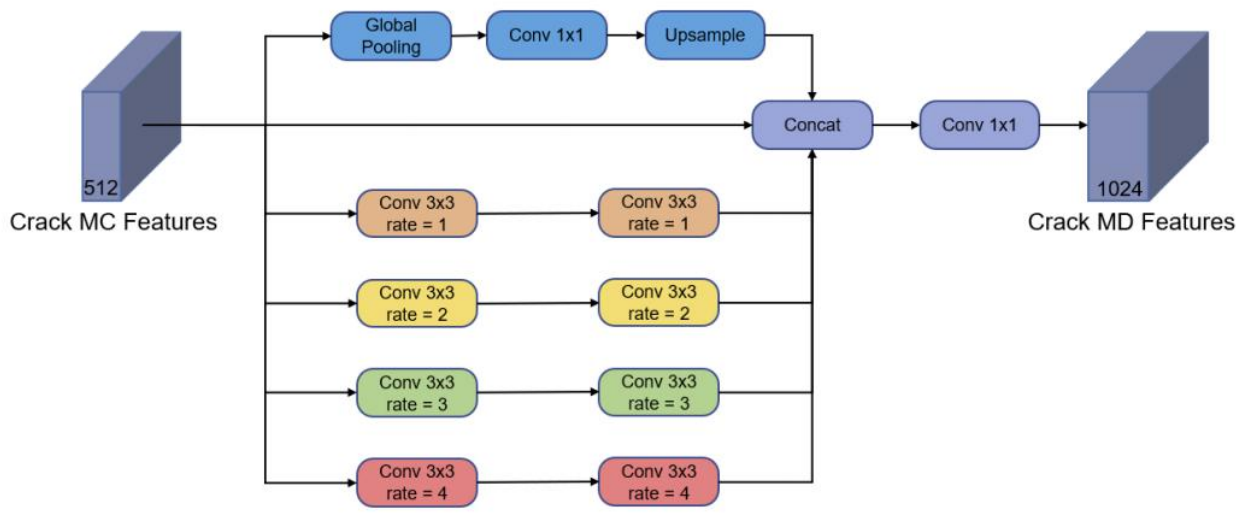

**Figure 10 Multi-Dilation (MD) module [17]**

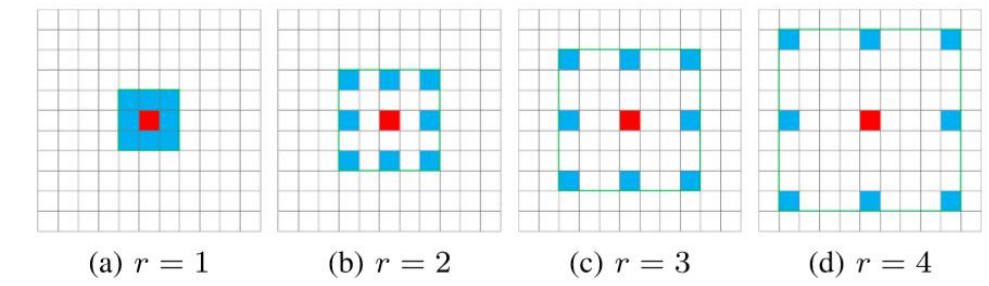

**Figure 11 Convolution kernels with different dilation rates**

[10] Source: https://keras.io/zh/ 2020-03-16

**Table 10 Test result and performance comparison of some backbone models**

| Method | | Accuracy | Time consuming (ms/img) (Image resolution) | |
|---|---|---|---|---|
| Data set | Network model | | Testing time | Predicting time (Image loading and pre-processing time included) |
| CCIC | TL-MobileNetV1 | **99.8%** | 0.9 ms/img (224x224) | / |
| Non-public, similar to CCIC | Self-built CNN [20] | 97.95% | / | 4500 ms/img (5888x3584) |
| Non-public, similar to CCIC | TL-VGG16 [13] | 90.0% | / | / |
| Non-public, similar to CCIC | TL-InceptionV3[14] | 97.4% | / | / |
| SDNET2018 | TL-MobileNetV1 | 94.4% | 1.0 ms/img (224x224) | **7.0** ms/img (224x224) |
| SDNET2018 | TL-MobileNetV2 | 94.8% | 1.0 ms/img (224x224) | 8.1 ms/img (224x224) |
| SDNET2018 | TL-InceptionV3 | **96.1%** | 1.7 ms/img (224x224) | **16.1** ms/img (224x224) |
| Non-public, similar to SDNET2018 | ED-FPCNet [17] | **97.5%** | / | **67.9** ms/img (288x288) |

Note: [1] As some of the crack data sets for benchmark test performance comparison are not disclosed (although they are very similar) and the computing platforms are somehow different, the performance comparison in this table is not completely strict and is only for reference. [2]TL is short for Transfer Learning; ED is short for Encoder-Decoder.

The MD module, as shown in Figure 12, the input are MD features and MC features, and the output is the optimized MD features after weighted fusion.

1) The SEU module first restores the resolution of the crack MD features through transposed convolution. Then,

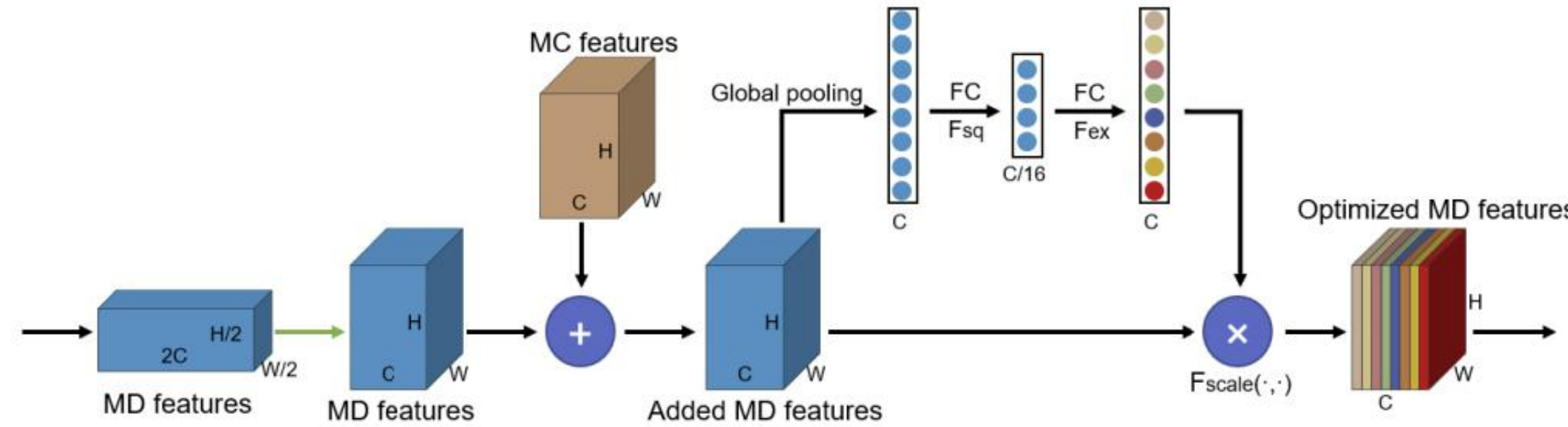

**Figure 12 SE-Upsampling module [17]**

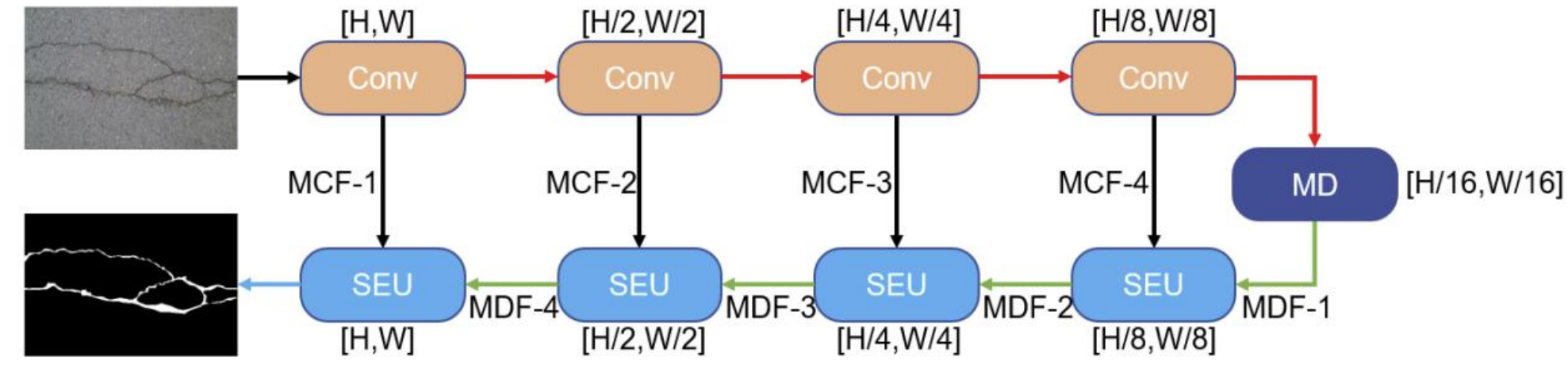

**Figure 13 Network architecture of FPCNet [17]**

it adds the MC features to the MD features in order to fuse the associated crack information concerning the edge, pattern, texture among others.

2)Subsequently, the Squeeze-and-Excitation (SE) operation is applied to the added MD features to learn the weights of different features. Global pooling is first performed to obtain the global information of the C channels. After squeeze ($F_{sq}$) and excitation ($F_{ex}$) (two fully-connected layers) of the global information, the weight of each feature for its channel is obtained. Through the SE learning, the SEU module can adaptively assign different weights to different crack features such as the edge, pattern, and texture.

3) Finally, each feature in the added MD features is multiplied ($F_{scale}$) by its corresponding weight to obtain the optimized MD features.

In Figure 12, the green arrow indicates the transposed convolution. H, W, and C represent the length, width, and number of channels of the features, respectively. The model architecture of FPCNet [17] is shown in Figure 13.

(3) Testing performance

The testing performance of crack detection on CFD data set are shown in Table 11. It can be seen that the precision, recall and F1 scores of FPCNet [17] are all ahead of machine learning, CNN, FCN and other methods, and it is the SOTA model. The prediction speed of FPCNet [17] is also relatively fast. Predicting a single 288×288 image sample on NVIDIA GTX1080Ti GPU takes only 67.9ms (i.e., 14.7 FPS). It can realize real-time detection on embedded platform.

(4) Identification effect

Comparison of detection effect of FPCNet [17] and Structure-Prediction CNN [8] etc. methods on CFD is shown in Figure 14.

**Table 11 Testing performance of crack detection on CFD**

| Method | Annotation error (pixel) | Precision | Recall | *F*1 |
|---|---|---|---|---|
| CrackForest | 5 | 0.8228 | 0.8944 | 0.8571 |
| MFCD | 5 | 0.8990 | 0.8947 | 0.8804 |
| Structured-Prediction CNN [8] | 2 | 0.9119 | 0.9481 | 0.9244 |
| FCN | 2 | 0.9729 | 0.9456 | 0.9590 |
| FPCNet [17] | 2 | **0.9748** | **0.9639** | **0.9693** |

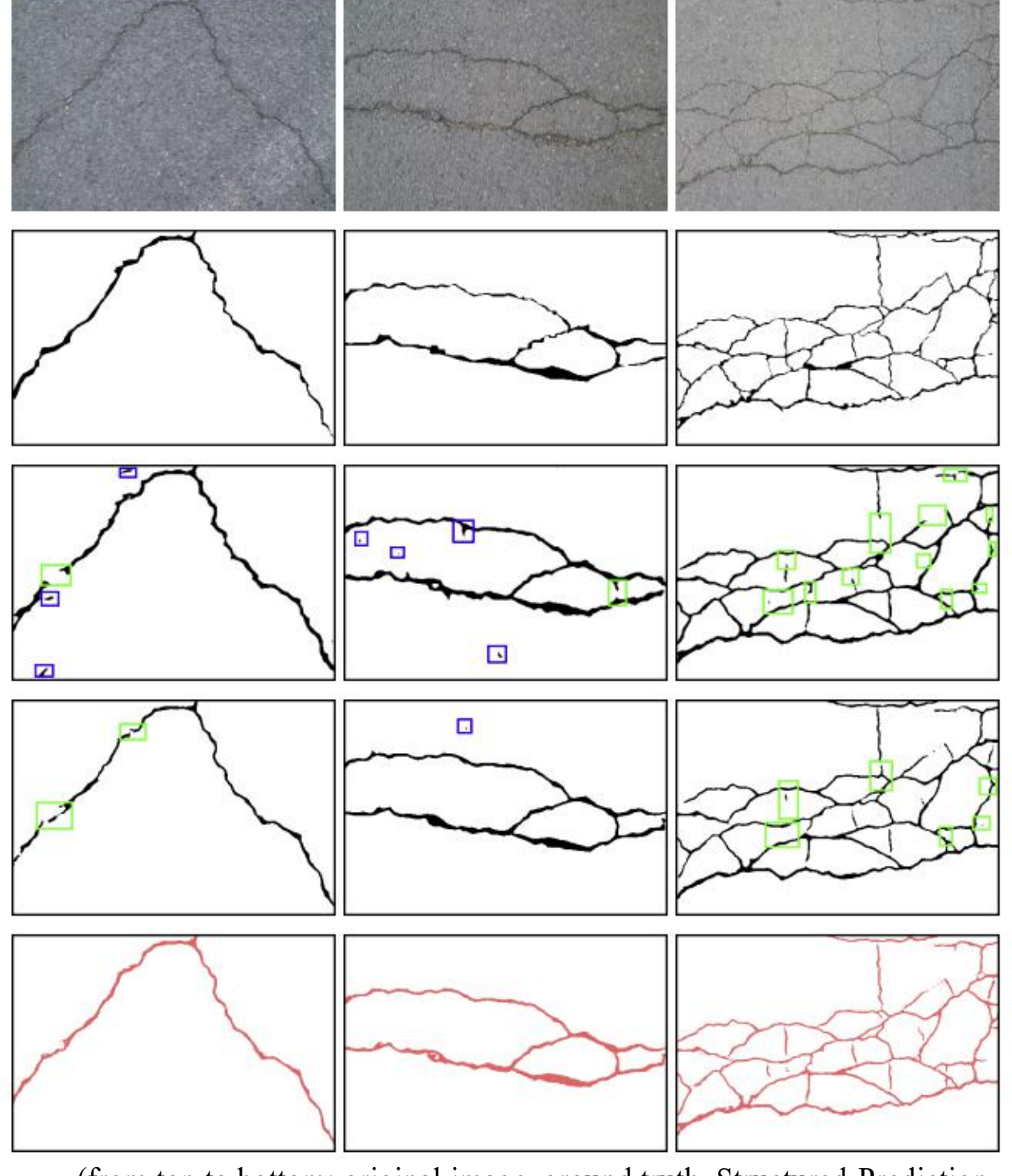

(from top to bottom: original image, ground truth, Structured-Prediction CNN，FCN，FPCNet)

**Figure 14 Comparison of detection effect between FPCNet [17] and Structured-Prediction CNN [8] etc. on CFD**

These two models, as well as CrackNet/CrackNet-V [9-10], can achieve pixel-level accuracy in crack identification (2-pixel annotation error). As intended, FPCNet [17] is more accurate than other methods in identifying the edges, textures and details of various complex forms of cracks, such as alligator cracks.

4. Summary and conclusion

Compared to NDT and health monitoring method for cracks in engineering structures, surface crack detection or identification based on visible light visual images is a non-contact one. It is not limited by the material of the tested object and is easy to achieve online real-time automation, thus, has the advantages of fast speed, low cost and high precision. Thus, it is expected to be applicable to preventive testing or monitoring scenarios such as routine patrol inspection, and defect detection of objects the geometric topological form of which can be regarded as 2D plane. If the input images or signals are acquired by acoustic emission etc. penetrating equipment, internal cracks may also be monitored or identified.

Firstly, typical pavement (concrete also) crack public data sets for classification, location and segmentation were collected, and the characteristics of sample images as well as the random variable factors, including environmental, noise and interference etc., were summarized. Subsequently, the advantages and disadvantages of three main crack identification methods (i.e., hand-crafted feature engineering, machine learning, deep learning) were compared. Finally, from the aspects of model architecture, testing performance and predicting effectiveness, the development and progress of typical deep learning models, including self-built CNN, transfer learning and encoder-decoder, which can be easily deployed on embedded platform, were reviewed and evaluated. From which, we can see the evolution of CNN model architecture, and the obvious improvement of crack identification performance and effectiveness due to computing power enhancement and algorithm optimization.

Currently, it has been able to realize real-time pixel-level crack identification and detection on embedded platform with a single deep learning model. For instance, the entire crack detection average time cost of an image sample is less than 100ms, either using the encoder-decoder method (i.e., FPCNet) or the transfer learning method based on InceptionV3. It can be reduced to less than 10ms with transfer learning method based on MobileNet (a lightweight backbone base network). In terms of accuracy, testing accuracy can reach over 99.8% on CCIC data set which is easily identified by human eyes. On SDNET2018 data set, some samples of which are difficult to be identified by human eyes, FPCNet can reach 97.5%, while transfer learning method is close to 96.1%. It is expected to further improve the accuracy with the ensemble of single model mentioned above.